# Multi-scale interaction network for stereo image super-resolution

Liyi Xu, Lin Qi*
Ocean University of China, Qingdao, 266100, China
*qilin@ouc.edu.cn

## ABSTRACT

Stereo image super-resolution aims to generate high-resolution images by leveraging complementary information from binocular systems. Although previous studies have achieved impressive results, the potential of intra-view and cross-view information has not been fully exploited. To address this issue, we propose a novel multi-scale interaction network for stereo image super-resolution. Specifically, we design a Multi-scale Spatial-Channel Attention Module that utilizes multi-scale large separable kernel attention and simple channel attention to improve intra-view feature extraction. Additionally, we propose a Dual-View Epipolar Attention Module, utilizing an optimal transport algorithm to achieve more accurate matching along the epipolar line. Extensive experimental and ablation studies show that our method achieves competitive results that outperform most SOTA methods.

## 1. INTRODUCTION

Binocular cameras and stereo images are widely used in autonomous driving and virtual reality (VR) scenarios[1]. However, the resolution and quality of stereo images are often limited by optical sensors and acquisition conditions, hindering their application in emerging fields. As a result, stereo image super-resolution techniques, which recover high-resolution (HR) stereo images from low-resolution (LR) counterparts without altering the stereo imaging system, have garnered significant attention from academia and industry. Stereo image super-resolution leverages not only the image information within a single view but also the overlapping regions between views to enhance super-resolution reconstruction.

Unlike single image super-resolution (SISR) methods, stereo image super-resolution (STSR) methods leverage both single-view image information and cross-view information from stereo images to enhance STSR. Existing frameworks typically mitigate these issues by integrating an extraction unit for intra-view details alongside a cross-view parallax attention mechanism (PAM). By doing so, they aim to simultaneously bolster the capabilities of internal feature processing and the integration of information between views. For instance, PASSRnet and iPASSRnet[2] utilize residual blocks for intra-view feature extraction and propose their respective parallax attention modules to enhance stereo image reconstruction by leveraging cross-view information along the epipolar line. Although these networks demonstrate improved performance using PAM, the effectiveness of one-stage PAM interaction heavily depends on the quality of the extracted features, potentially leading to the neglect of additional useful cross-view information. To address this limitation, some methods have implemented multi-stage PAM interactions within their networks. NAFSSR employs nonlinear activation-free blocks from NAFNet[3] for intra-view feature extraction and introduces the Stereo Cross Attention Module as a cross-view fusion block, stacking these blocks to achieve improved performance over one-stage PAM. In a similar vein, SwinFSR employs cascaded Swin Transformer layers to handle single-view feature extraction, while incorporating Residual Cross Attention Modules for the integration of cross-view data. Although multi-stage PAM techniques demonstrate enhanced metrics, they are constrained by the use of feature extractors with invariant receptive fields during the generation of attention maps. This approach is inflexible and limits the ability to fully exploit features, as both surrounding and remote pixels play crucial roles in reconstruction. Furthermore, these methods consider normalization in only one direction when generating the attention map, potentially leading to suboptimal matches.

To address these challenges, we propose an innovative Multi-scale Interaction Network for Stereo Image Super-Resolution. Specifically, to extract intra-view features across multiple scales, we introduce a multi-scale spatial-channel attention module that leverages large separable kernel attention, enhancing the model's capacity to capture fine details. Furthermore, we apply the optimal transport algorithm to normalize the original cost matrix along both rows and columns, achieving a more reliable match on the epipolar line. This approach has proven effective in related tasks such as sparse feature matching and stereo matching.

## 2. METHODOLOGY

To reconstruct high-quality stereo images, this paper proposes a new Multi-scale Interaction Network for stereo image super-resolution. As shown in Figure 1., the shallow feature extraction module first applies a 3×3 convolution to extract shallow feature embeddings from LR images. Next, N stacked Multi-scale Spatial-Channel Attention Blocks (MSCAB) are employed to further refine intra-view features. To facilitate interaction and fusion of information between the two views, the Dual-view Epipolar Attention Module (DEAM) is inserted after the MSCAB blocks. This module uses stereo features from the previous intra-view extraction stage to perform cross-view interactions, producing interactive features. In the image reconstruction module, a 3×3 convolution followed by a sub-pixel layer is used for upsampling, recovering a clear high-resolution stereo image from the extracted features.

### 2.1 Multi-scale spatial-channel attention block

Many existing stereo image super-resolution (SR) methods primarily utilize attention mechanisms, such as channel attention (CA) and self-attention (SA), to capture more informative features. However, these methods struggle to simultaneously capture both local information and long-range dependencies, often relying on attention maps with a fixed receptive field. Inspired by recent advances in visual attention research, which achieve comparable performance while maintaining computational efficiency even with larger kernel sizes, we propose a Multi-scale Spatial-Channel Attention Module (MSCAB) to address these limitations.

As shown in Figure 2., MSCAB consists of two components: the Multi-scale Spatial-Channel Attention Module (MSCAM) and the Simple FeedForward Network (SFFN). MSCAM integrates channel attention with multi-scale large separable kernel attention to capture both channel-specific and multi-scale spatial information. Channel attention computes global statistics of feature maps to prioritize the most significant channels. Simultaneously, multi-scale large separable kernel attention[4] employs kernels of varying scales to extract features at different scales, capturing long-range dependencies within the image. This enhances the model's ability to focus on multi-scale local information.

Combining these two attention mechanisms allows MSCAM to effectively extract multi-scale information from the input image, leading to more accurate texture detail recovery. Following the architecture of NAFNet, we include a Simple FeedForward Network (SFFN) after MSCAM to extract deeper features. It is important to note that MSCAM and SFFN do not use nonlinear activation functions. Instead, they employ a Simple Gate operation, performing pixel-wise multiplication along the channel dimension. This operation, also known as the star operation in starnet[5], allows the generation of high-dimensional features while maintaining computation in low-dimensional space.

### 2.2 Dual-view epipolar attention module

Exploiting cross-view correlations and symmetric complementary cues is fundamental to stereo image super-resolution. Prevailing methods typically employ disparity attention for feature alignment. By enforcing epipolar constraints to restrict search space to 1D, this mechanism quantifies affinities via matrix multiplication and normalizes them into transfer weights using Softmax. These weights then guide the aggregation of auxiliary features to the reference view. However, this alignment remains imperfect due to intrinsic cross-view matching errors. To address cross-view feature interaction, this chapter introduces the Optimal Transport (OT) model as a substitute for traditional Softmax-based attention computation. By imposing dual constraints on the row and column directions of the cost matrix, OT theoretically guarantees that the generated attention maps strictly adhere to epipolar geometry constraints, thereby achieving an optimal assignment of pixel-wise matching information[6]. To derive the optimal coupling matrix $T$ , entropy-regularized optimal transport addresses a specific optimization formulation. This process assumes the existence of a cost matrix $M$ linking two marginal distributions ( $a$ and $b$ ) of dimension $n$ :

$$
\begin{aligned}
&T = \underset{T \in R_{+}^{I_w \times I_w}}{\arg\min} \sum_{i,j=1}^{I_w, I_w} T_{ij} M_{ij} - \gamma E(T) \\
&s.t. T 1_{I_w} = 1, T^{\top} 1_{I_w} = 1
\end{aligned} \tag{1}
$$

In this formulation, E(T) represents the regularization of entropy T effectively acts as the solution to the standard assignment problem. As illustrated in Figure 2, the input stereo features $X_L$ and $X_R$ (each $\in \mathbb{R}^{H \times W \times C}$ ) undergo layer normalization and convolution operations to produce the features $U_L$ and $U_R$ (also $\in \mathbb{R}^{H \times W \times C}$ ). The cost matrix M can be obtained by

$$M = U_L U_R / \sqrt{C} \tag{2}$$

Here, C represents the number of channels in the features $U_L$ and $U_R$. The optimal attention map T is obtained using the iterative Sinkhorn algorithm, as described in Algorithm 1. The Sinkhorn algorithm incurs low computational overhead as it computes the assignment map using a finite number of iterations. In the proposed method, the number of iterations is set to 10. Then, the cross-view fusion features can be obtained by

$$\begin{aligned} F_{R\to L} &= TV_R, F_{L\to R} = T^{\top} V_L \\ F_L &= \gamma_L F_{R\to L} + X_L, F_R = \gamma_R F_{L\to R} + X_R \end{aligned} \tag{3}$$

Where $V_L$ and $V_R$ are the features of $X_L$ and $X_R$ obtained through a $1\times1$ convolution operation. The interacted cross-view information $F_{R\to L}$ and $F_{L\to R}$, along with the intra-view information $X_L$ and $X_R$, are fused via element-wise addition. The trainable channel scales $\gamma_L$ and $\gamma_R$ are initialized to zero to stabilize training, ensuring that the model learns these parameters effectively over time.

Algorithm 1. The implementation details of Sinkhorn Normalization.

Input: raw cost matrix $M \in R^{H\times W\times W}$, $\log \mu \in \{1/(2W)\}^{H\times 1\times W}$, $\log v \in \{1/2W\}^{H\times W\times 1}$, iters
Output: Updated attention weights
Initialize $u = 0$ and $v = 0$, with the same shape as $\log \mu$ and $\log v$;
For idx = 1 to iters do:
$v = \log v - \log sum \exp(M + u \otimes 1_3, \dim = 2);$
$u = \log \mu - \log sum \exp(M + v \otimes 1_2, \dim = 3);$
$T = \exp(M + u \otimes 1_3 + v \otimes 1_2 + \log 2W)$
return $T$

### 2.3 Loss Function

Recognizing that the core challenge of image super-resolution is the retrieval of high-frequency nuances, we leverage a combination of loss constraints from both the spatial and frequency domains to supervise texture reconstruction. Formally, the system accepts a pair of low-resolution stereo inputs with the aim of synthesizing their high-resolution equivalents. The optimization objective used to drive this process is defined as:

$$L_{total} = \frac{1}{N}\sum_{i=1}^{N} || I_{HR}^{L,R} - I_{SR}^{L,R} ||^2 + \lambda \frac{1}{N}\sum_{i=1}^{N} || FFT(I_{HR}^{L,R}) - FFT(I_{SR}^{L,R}) || \tag{4}$$

Where $\lambda$ is set to 0.01, and $FFT(\cdot)$ denotes the Fourier transform.

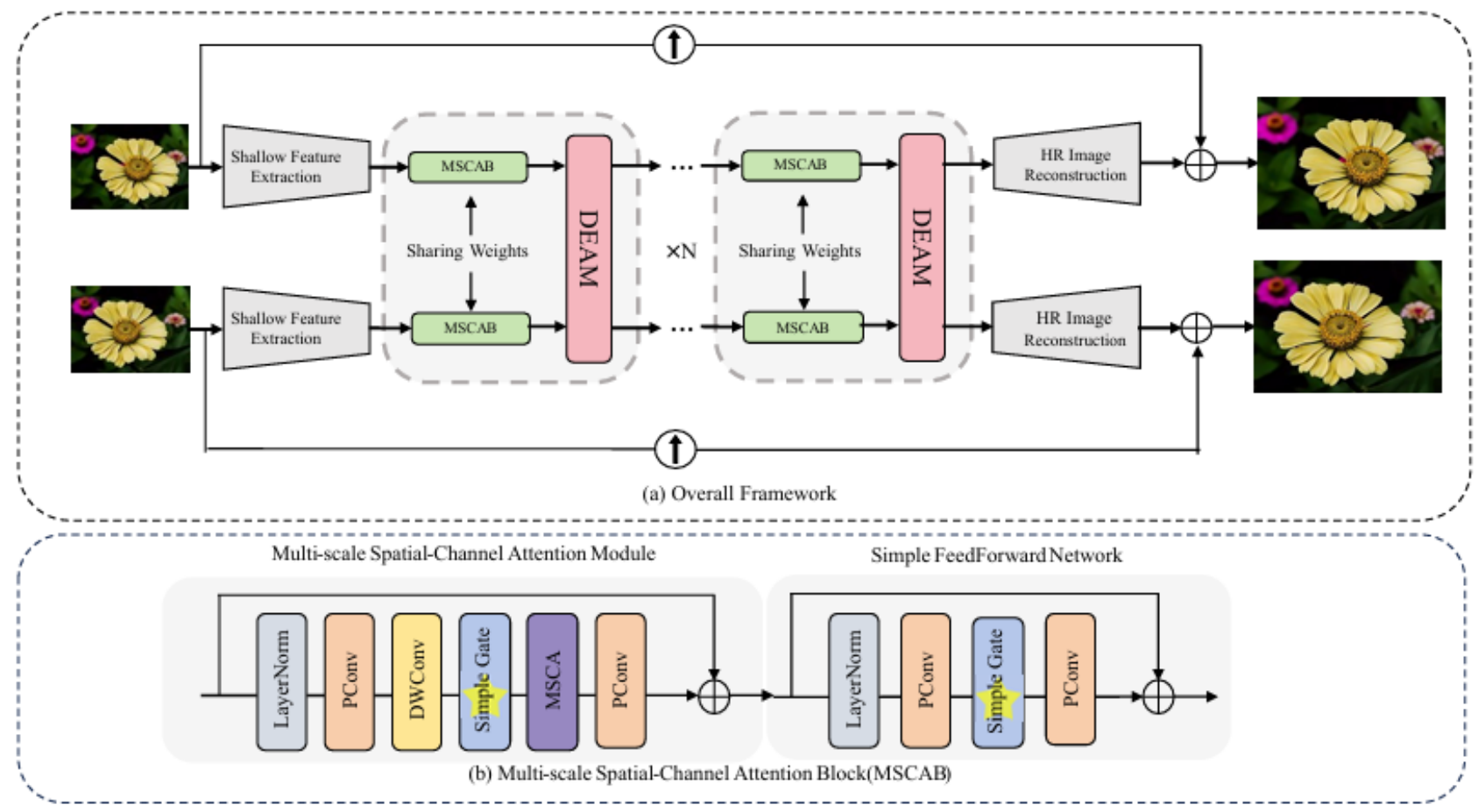

Figure 1. An overview of our MSINet.

## 3. EXPERIMENTS

### 3.1 Experimental settings

Following the previous approach, we use 800 stereo images from Flickr1024 training dataset and 60 stereo images from Middlebury dataset as training data. For testing, we use four benchmark datasets: KITTI2012, KITTI2015, Middlebury, and Flickr1024 test sets. The number of MSINet blocks is set to 32. For the optimization process, we employ the Lion optimizer[7] with the weight decay configured to 0. The learning rate is initialized at $3\times10^{-4}$ and subsequently decays to $1\times10^{-8}$ following a cosine annealing strategy. All experimental configurations and dataset processing follow prior studies.

### 3.2 Comparison with the state-of-the-art

As shown in Table 1, Compared with SISR methods EDSR, RCAN and stereo image SR methods, iPASSRnet, SSRDE-FNet[8], NAFSSR, SwinFSR[9] and MSSFNet[10], our proposed MSINet achieves significant PSNR and SSIM. Moreover, the qualitative evaluation in Figure 4 demonstrates that the details recovered by our method are clearer and more closely resemble the groundtruth images.

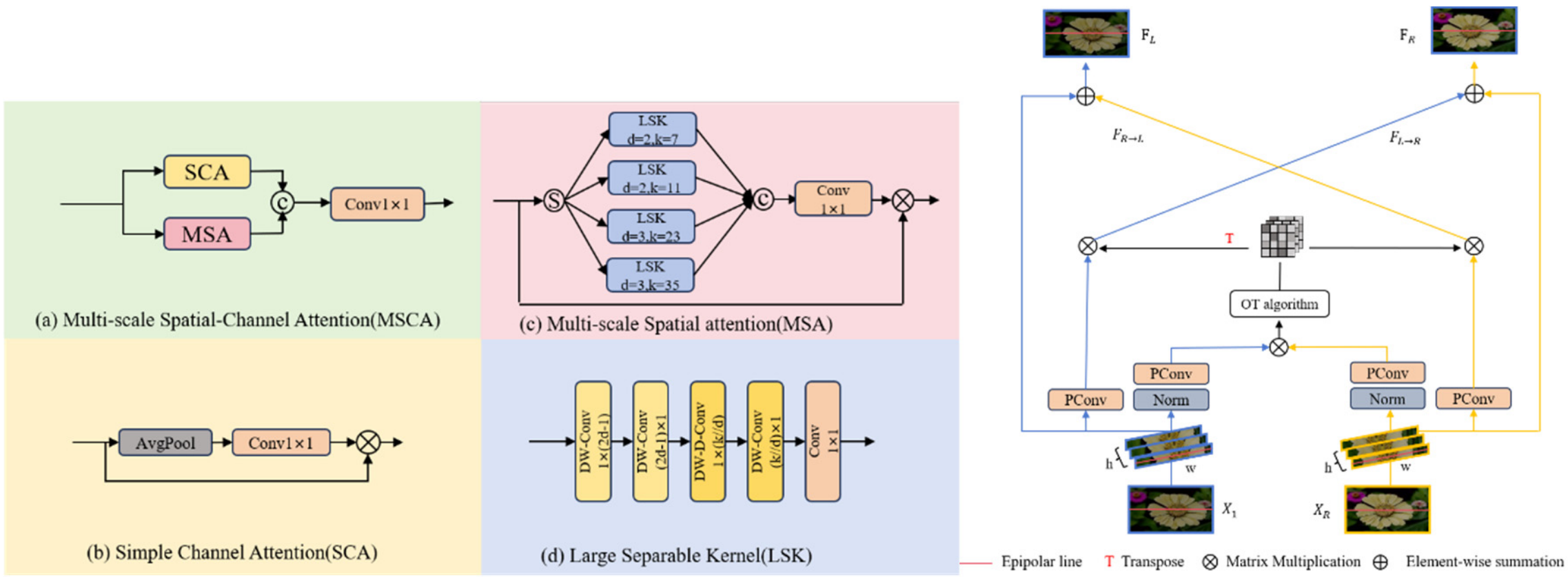

Figure 2. Multi-scale Spatial-Channel Attention Block(left) and Dual-view Epipolar Attention Module(right).

To further investigate the reconstruction quality of super-resolution methods in occluded and texture-less regions, we selected 20 images from the synthetic SceneFlow dataset. These images were downsampled by a factor of 4, and various STSR models were subsequently applied to perform 4 times super-resolution reconstruction, restoring the original resolution for stereo matching. The RAFT stereo matching model was employed for this experiment. The visualization and performance comparison results are presented in Figure 4, which illustrates the stereo matching visualizations and the End-Point Error (EPE). As observed, our method demonstrates superior matching performance and accuracy. And see in figure 3.

### 3.3 Ablation study

To evaluate the effectiveness of our method, we initially conducted ablation studies by removing all additional modules and then compared the performance of the SCAM module and the DEAM module. Table 2 shows that our DEAM interaction module outperforms SCAM module. The inclusion of multi-scale spatial attention within the MSCAM module enhances the model's performance, increasing the PSNR by 0.08 dB. Furthermore, incorporating the frequency domain loss function sharpens the model's focus on high-frequency details, leading to a 0.04 dB improvement in PSNR.

Table 1. Quantitative comparisons for ×4 SR with PSNR/SSIM values. Higher PSNR and SSIM mean better performance.

| Method | Params | KITTI2012 | KITTI2015 | Middlebury | Flickr1024 |
|---|---|---|---|---|---|
| EDSR | 38.9M | 26.35/0.8015 | 26.04/0.8039 | 29.23/0.8397 | 23.46/0.7285 |
| RCAN | 15.4M | 26.44/0.8029 | 26.22/0.8068 | 29.30/0.8397 | 23.48/0.7286 |
| iPASSRnet | 1.42M | 26.56/0.8053 | 26.32/0.8084 | 29.16/0.8367 | 23.44/0.7287 |
| SSRDE-FNet | 2.24M | 26.70/0.8082 | 26.45/0.8118 | 29.38/0.8411 | 23.59/0.7352 |
| NAFSSR | 1.56M | 26.93/0.8145 | 26.76/0.8203 | 29.72/0.8490 | 23.88/0.7468 |
| MSSFNet | 1.82M | 26.97/0.8158 | 26.82/0.8219 | 29.77/0.8502 | 23.99/0.7508 |
| SwinFSR | 9.76M | 27.03/0.8143 | 26.83/0.8213 | **32.45/0.8891** | 23.83/0.7471 |
| MSINet(ours) | 2.04M | **27.07/0.8195** | **26.88/0.8253** | 29.98/0.8547 | **24.03/0.7541** |

Table 2. Ablation studies on Flickr1024 Dataset, ×4 SR.

| DEAM | SCAM | MSCAM | Frequency Loss | PSNR/SSIM |
|---|---|---|---|---|
| × | × | × | × | 23.57/0.7312 |
| × | √ | × | × | 23.84/0.7452 |
| √ | × | × | × | 23.85/0.7457 |
| √ | × | √ | × | 23.93/0.7494 |
| √ | × | √ | √ | 23.97/0.7518 |

Left
Right
EDSR iPASSR SSRDEFNet NAFSSR Ours GT

Figure 3. Visual quality assessment of 4 upscaling results generated by competing approaches on the Flickr1024 set.

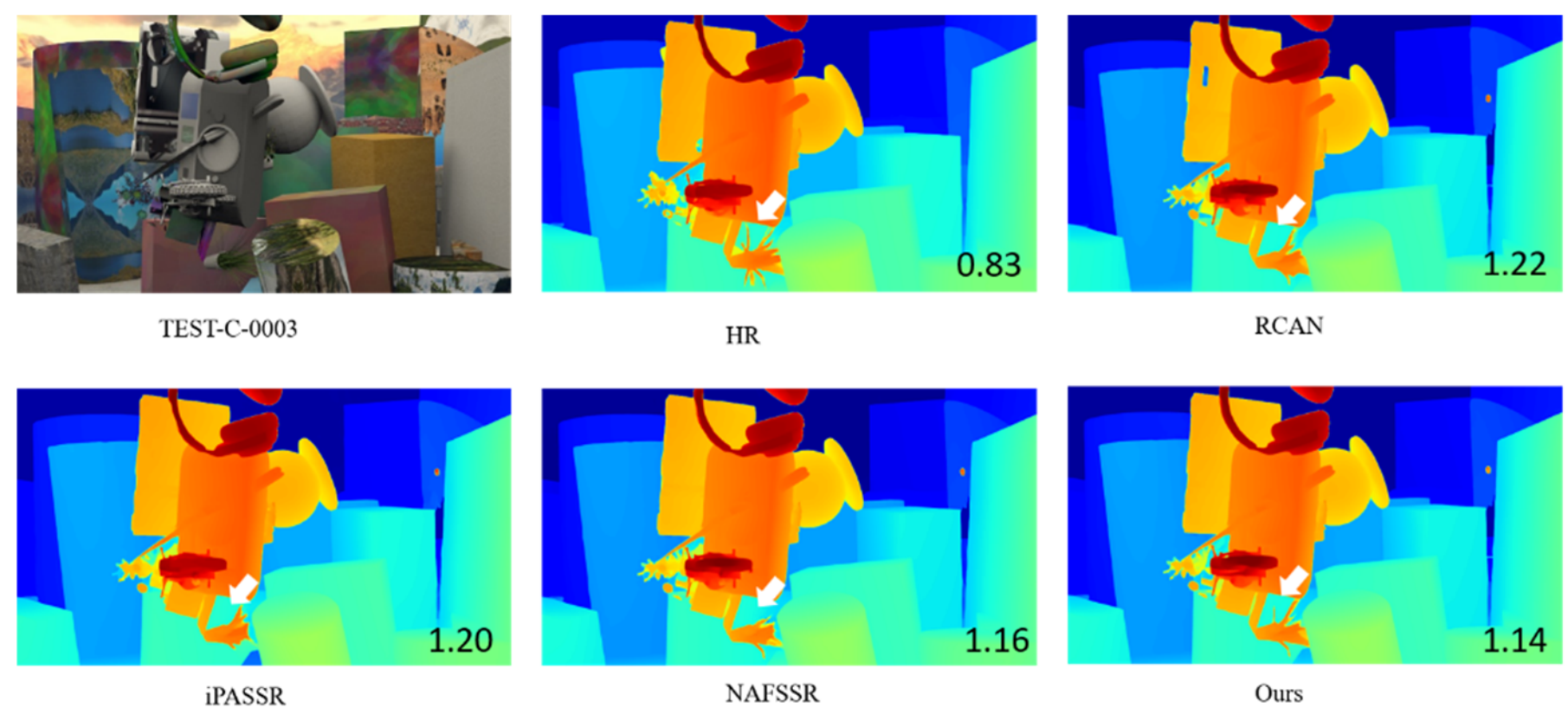

Figure 4. Visual comparison for disparity maps obtained from 4× SR stereo images on sceneflow test set.

## 4. CONCLUSION

In this paper, we propose MSINet, an efficient stereo image super-resolution method that leverages multi-scale spatial attention with large kernels and the optimal transport algorithm. MSINet employs the MSCAM and DEAM modules to effectively extract single-view information and fuse cross-view features. The MSCAM module uses large kernel separable convolutions with varying receptive fields to capture multi-scale information, while the DEAM module applies the optimal transport algorithm to enhance feature map alignment. Extensive experimental evaluations show that MSINet outperforms leading state-of-the-art models in stereo image super-resolution, achieving superior PSNR and SSIM scores.